\pdfoutput=1

\documentclass[11pt]{article}

\usepackage{acl}

\usepackage{times}
\usepackage{latexsym}
\usepackage{graphicx}
\usepackage{multirow}
\usepackage{xcolor,colortbl}
\usepackage{dsfont}
\usepackage{amsmath}
\usepackage{amssymb}
\usepackage{listings}

\definecolor{codegreen}{rgb}{0,0.6,0}
\definecolor{codegray}{rgb}{0.5,0.5,0.5}
\definecolor{codepurple}{rgb}{0.58,0,0.82}
\definecolor{backcolour}{rgb}{0.95,0.95,1}

\lstdefinestyle{json}{
    backgroundcolor=\color{white},   
    basicstyle=\fontsize{8}{9}\tt,
    keywordstyle=\color{codepurple},
    numberstyle=\tiny\color{codegray},
    breakatwhitespace=false,         
    breaklines=true,                 
    captionpos=b,                    
    keepspaces=true,                 
    numbers=left,
    numbersep=5pt,                  
    showspaces=false,                
    showstringspaces=false,
    showtabs=false,                  
    tabsize=2,
    linewidth=\textwidth,
    frame=single
}

\usepackage{tcolorbox}
\tcbuselibrary{listings,breakable}
\tcbset{colframe=black,colback=white,size=small,colbacktitle=gray!30!white,coltitle=black,fonttitle=\bfseries,fontupper=\footnotesize\ttfamily}

\usepackage{pgfplots}
\pgfplotsset{compat=1.16}
\usepackage{tikz}
\usepgfplotslibrary{groupplots}

\definecolor{Gray}{gray}{0.85}
\definecolor{LightCyan}{rgb}{0.88,1,1}
\definecolor{White}{rgb}{1.0, 1.0, 1.0}

\newcolumntype{a}{>{\columncolor{Gray}}r}

\usepackage[T1]{fontenc}

\usepackage[utf8]{inputenc}

\usepackage{microtype}
\usepackage{booktabs}

\usepackage{inconsolata}

\usepackage[strict]{changepage}
\definecolor{darkblue}{rgb}{0.0, 0.0, 0.55}
\usepackage{framed}
\definecolor{formalshade}{rgb}{0.95,0.95,1}
\newenvironment{formal}{%
  \MakeFramed{\advance\hsize-\width\FrameRestore}%
  \noindent\hspace{-4.55pt}
  \begin{adjustwidth}{}{7pt}%
  \vspace{2pt}\vspace{2pt}%
}
{%
  \vspace{2pt}\end{adjustwidth}\endMakeFramed%
}

\title{Propagation and Pitfalls: Reasoning-based Assessment of \\Knowledge Editing through Counterfactual Tasks}

\author{
     Wenyue Hua$^{\ddagger}$$^*$ , Jiang Guo$^\dagger$$^*$ , Mingwen Dong$^\dagger$, Henghui Zhu$^\dagger$, Patrick Ng$^\dagger$, Zhiguo Wang$^\dagger$\\
    $^\ddagger$Rutgers University, New Brunswick~~~~
    $^\dagger$AWS AI Labs\\
    wenyue.hua@rutgers.edu,\\
    \{gujiang, mingwd, henghui, patricng, zhiguow\}@amazon.com
}

\begin{document}                                   
\maketitle
\renewcommand{\thefootnote}{\fnsymbol{footnote}}
\footnotetext[1]{Equal contribution. Work done while Wenyue Hua was an intern at AWS AI Labs.}
\renewcommand{\thefootnote}{\arabic{footnote}}

\begin{abstract}
Current approaches of knowledge editing struggle to effectively propagate updates to interconnected facts.
In this work, we delve into the barriers that hinder the appropriate propagation of updated knowledge within these models for accurate reasoning. 
To support our analysis, we introduce a novel reasoning-based benchmark -- \textbf{ReCoE} (\textbf{Re}asoning-based \textbf{Co}unterfactual \textbf{E}diting dataset) -- which covers six common reasoning schemes in real world.
We conduct a thorough analysis of existing knowledge editing techniques, including input-augmentation, finetuning, and locate-and-edit.
We found that all model editing methods show notably low performance on this dataset, especially in certain reasoning schemes.
Our analysis over the chain-of-thought generation of edited models further uncover key reasons behind the inadequacy of existing knowledge editing methods from a reasoning standpoint, involving aspects on \textit{fact-wise editing}, \textit{fact recall} ability, and \textit{coherence} in generation. We will make our benchmark publicly available.
\end{abstract}

\section{Introduction}
Contemporary language models demonstrate a remarkable capacity to encode extensive factual information, rendering them highly useful as a knowledge base for real-world applications. Yet, the challenge of rapidly outdated knowledge persists, giving rise to a wide range of methods for knowledge updating, such as in-context learning \cite{vu2023freshllms}, continual pretraining \cite{constraint}, locate-and-edit \cite{ROME,MEMIT}, and meta-learning \cite{MEND}.

\begin{figure}[htbp]
    \centering
    \includegraphics[width=77mm]{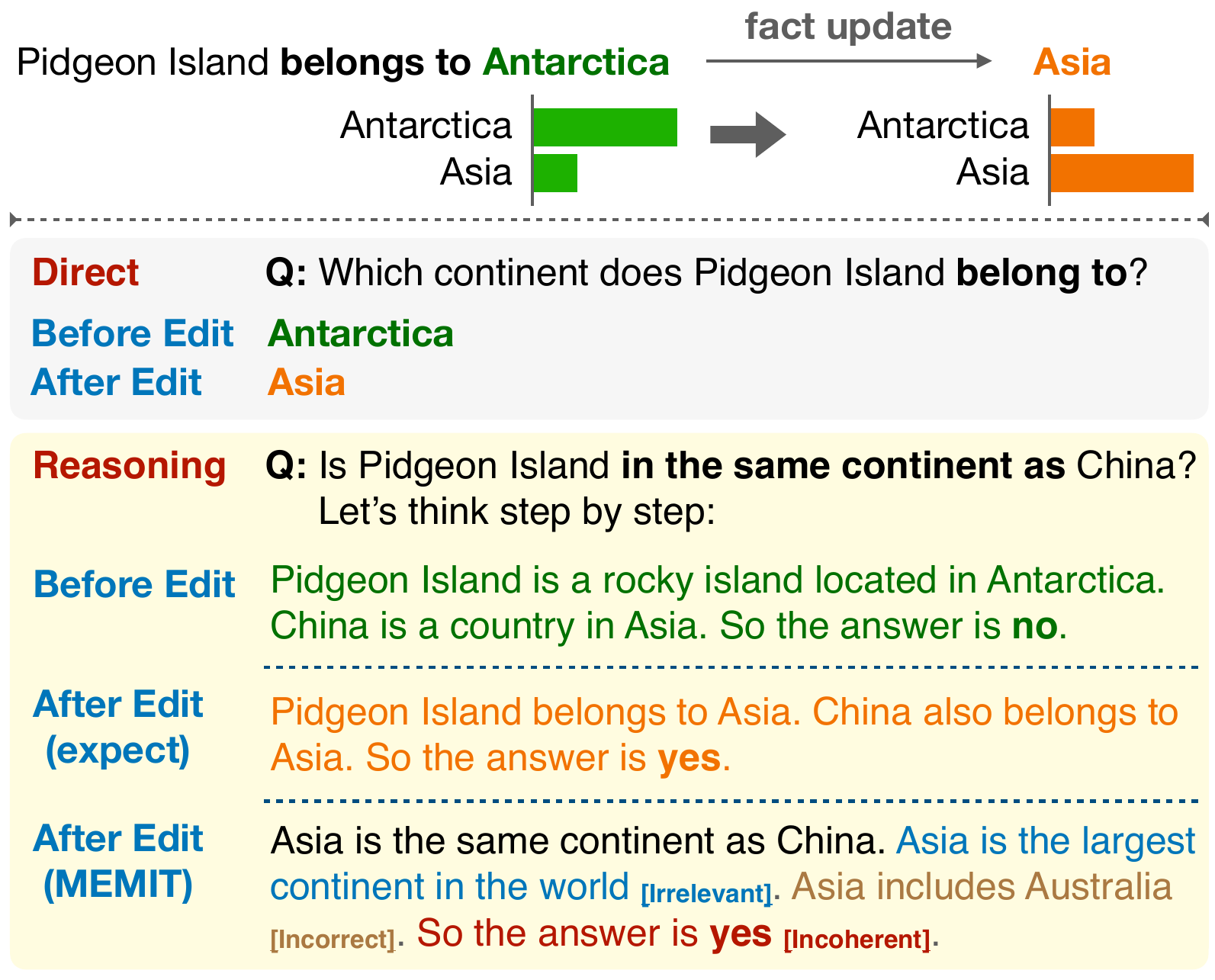}
    \caption{An example of reasoning-based assessment for knowledge editing. Existing methods perform well at answering the question of the edited fact, but fail on reasoning with the edited fact.}
    \vspace{-1em}
    \label{fig:illustration}
\end{figure}


Despite the success of fact-wise editing, recent studies \cite{MQuAKE, entityedit, pinter2023emptying} show that current model editing methods struggle to effectively propagate updates to interconnected facts.
However, the reason behind the ineffective knowledge propagation remains largely unexplored. Our observation in fact-editing experiments indicates that models respond unexpectedly post-editing. For instance, upon applying MEMIT \cite{MEMIT} for fact editing, the model was unable to reliably recall pertinent edited information and produces an incoherent chain-of-thought (CoT) \cite{chainofthought} for answering the question, as demonstrated in the MEMIT-based generation example in Figure \ref{fig:illustration}.

In this work, we undertake an in-depth analysis of this phenomenon, concentrating on three essential competencies necessary for knowledge propagation on reasoning questions after model editing. We place special emphasis on analyzing the results of CoT prompting, which provides explicit reasoning steps that facilitate easier examination. Specifically, in the proposed analytical framework, we measure: (1) effectiveness of editing individual facts, (2) accuracy in recalling relevant facts, and (3) logical coherence of the thought process. 

To facilitate our investigation, we introduce a \textbf{Re}asoning-based \textbf{Co}unterfactual \textbf{E}diting dataset -- \textbf{ReCoE}, which covers six different reasoning schemes: \textit{superlative}, \textit{comparative}, \textit{sorting}, \textit{counting}, \textit{aggregation}, and \textit{subtraction}.
This dataset is designed to more accurately capture the complexities inherent in fact editing tasks.
In contrast to existing knowledge editing benchmarks which predominantly rely on synthetic data \cite{MQuAKE,ROME}, ReCoE incorporates reasoning questions that exhibit a higher degree of naturalness and more accurately reflect the reasoning-based queries encountered in real-world scenarios.


We first explored input-augmentation, an approach where new facts are added (prepended) only in-context, as an informal upper bound of model editing methods.
We then examined model editing methods including finetuning and MEMIT on the Tülu series \cite{tulu}, which are Llama-based instruction-tuned models of varying sizes.
Results show that all model editing methods achieve notably low performance on the ReCoE benchmark, especially in certain reasoning schemes, with scores close to zero.
Our analysis further unravels the effect of various knowledge editing methods on the reasoning abilities of language models.
We demonstrate that all editing methods result in a significant reduction in fact recall, indicating a key obstacle in effective utilization of the edited knowledge.
Surprisingly, models edited through locate-and-edit methods (i.e., MEMIT) exhibit a severe decline in their generation coherence, leading to nonsensical outputs, which suggests a substantial deterioration in their fundamental language modeling abilities.
We summarize our contributions of the paper as follows:
\begin{itemize}
    \setlength\itemsep{0em}
    \item We introduce a reasoning-based framework of assessing knowledge editing methods, covering key aspects that enables effective reasoning. Our analysis uncovers essential insights regarding the challenges and limitations associated with knowledge propagation.
    \item We introduce ReCoE, a novel yet challenging reasoning-based counterfactual editing benchmark covering a diverse set of reasoning schemes centered on real-world scenarios.
\end{itemize}

\begin{figure*}
    \centering
    \includegraphics[scale=0.43]{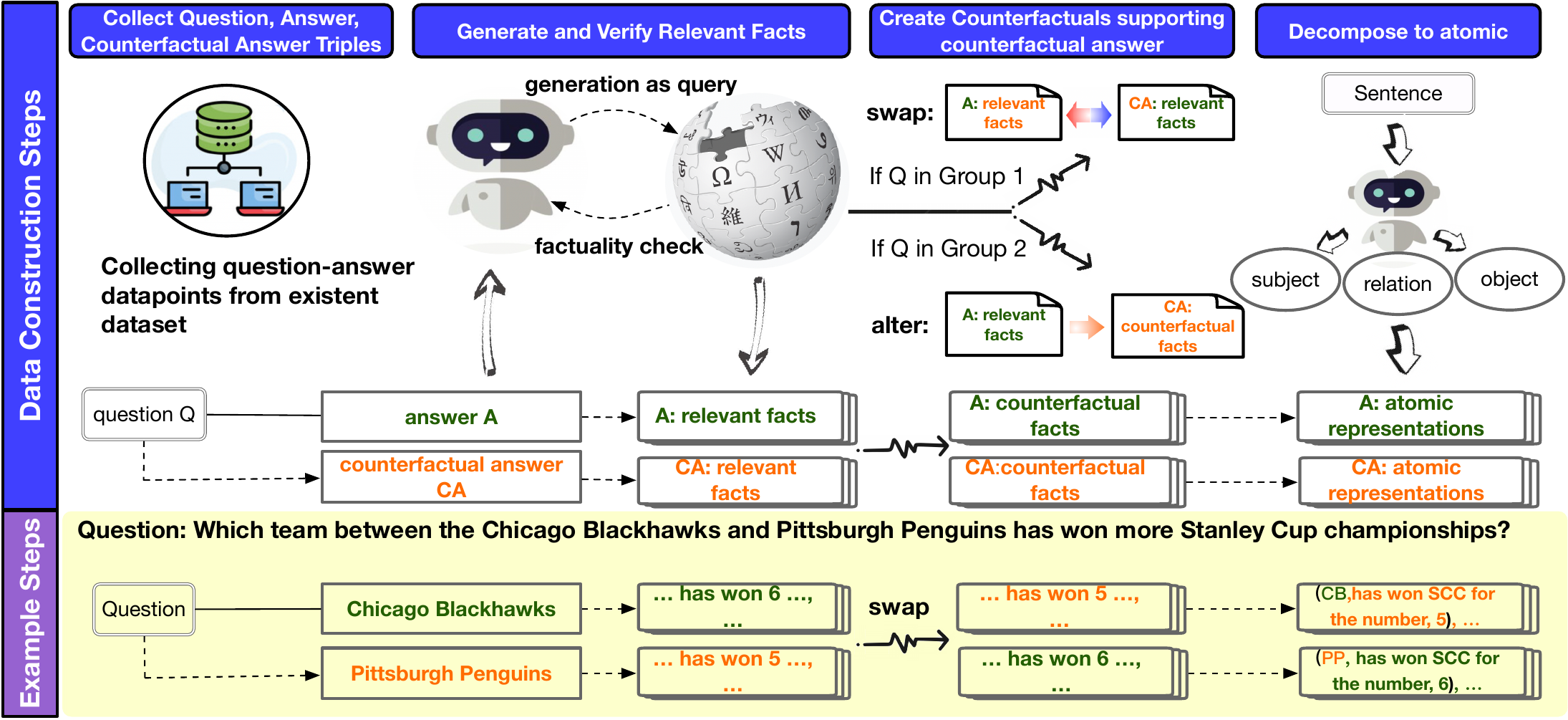}
    \caption{Step-by-step demonstration on dataset construction. There are three different types of lines in the diagram: (1) straight lines represent data sourced from existing datasets; (2) dashed lines denote data derived from Claude-generation processes; (3) zigzag lines symbolize data obtained through the corruption of other data. Group 1 includes superlative, comparative, and sorting questions, where we use ``swapping'' to create counterfactual facts. Group 2 represents counting, aggregation, and subtraction questions, where we use ``altering'' to create counterfactual facts. During the editing phase, data is processed either in text form or atomic triplets, contingent upon the specific requirements of the editor.}
    \label{fig:construction_pic}
\end{figure*}

\section{Related Work}\label{related}
\subsection{Model Editing Methods}
Existing model editing methods generally fall into four main categories \cite{wang2023knowledge}.

\paragraph{Finetuning-based methods} These techniques further finetune the model on new knowledge while minimizing the change in models and catastrophic forgetting. Examples of finetuning-based methods include \cite{constraint, chen2020recall, zhu2020modifying}.

\paragraph{Machine learning framing methods} These approaches treat the editing as a machine learning challenge. They learn hypernetworks (optimizers) to process model gradients. The goal is to produce an updated model that offers the desired output for the edited point while ensuring minimal prediction changes for other data points. Notable methods include MEND \cite{MEND}, KnowledgeEditor \cite{knowledgeedit}, SLAG \cite{slag}, and CaMeLS\cite{hu2023meta}.

\paragraph{Interpretability-centric methods} These methods focus on model interpretability. The objective is to pinpoint specific layers and parameters that primarily function for knowledge storage \cite{dai2021knowledge}. Once identified, these parameters are then edited, viewing them as linear associative memory storage units. ROME and MEMIT \cite{ROME,MEMIT} are prominent examples.

\paragraph{Retrieval-augmented methods} These techniques leverage retrieval-augmentation to update knowledge in prompting \cite{vu2023freshllms}. SERAC \cite{serac} and MeLLo \cite{MQuAKE} store new knowledge in memory. When relevant queries arise, they retrieve the pertinent knowledge from this storage, employing input augmentation to adjust the response. 

\subsection{Model Editing Benchmarks}
Several benchmarks have been introduced to assess the efficacy of model editing. \citet{ROME} introduced the COUNTERFACT dataset, specifically designed to evaluate the successful incorporation of counterfactual knowledge. This evaluation is segmented into three main criteria: (1) Efficacy determines if a particular piece of knowledge has been successfully integrated into the model, typically measured by conditional generation that mirrors the exact text used for editing (2) Paraphrase assesses the model's capability to generalize to paraphrased versions of the editing text, ensuring the retention of the original knowledge \cite{ROME} (3) Specificity ensures that the model remains unchanged with respect to irrelevant knowledge. There are many other datasets including Zero-Shot Relation Extraction (zsRE) \cite{MEND}, WikiGen \cite{MEND}, T-REx-100 \& T-REx-1000 \cite{elsahar2018t, dong2022calibrating}, MMEdit \cite{cheng2023can} (multi-modal model editing), \emph{etc}.

To evaluate knowledge propagation, \citet{entityedit} and \citet{MQuAKE} introduced ECBD and MQuAKE benchmarks respectively. ECBD measures the perplexity of a passage relevant to target knowledge (entity-relevant) before and after editing. Though it presents the difficulty of knowledge propagation, the context it evaluates on has a non-deterministic connection to the edited knowledge. MQuAKE employs multi-hop QA questions to gauge the model's accuracy after editing part or all of its reasoning components. However, it exclusively encompasses compositional questions generated from ChatGPT wherein the precise segments of knowledge required for effective propagation are overtly articulated within the question. Such format may not necessarily mirror real-world scenarios where the reasoning component could be implicit from the question.

In our research, we focus on factual knowledge editing and tackle the prevailing limitations on propagation observed in contemporary benchmarks. Our benchmark ReCoE incorporates a diverse set of 6 reasoning schemes, featuring more organic queries that mirror real-world scenarios. Additionally, we employ a reasoning-based framework to elucidate the underlying challenges of knowledge propagation. Drawing from our discoveries, we aspire to provide valuable insights that will guide and shape the future trajectories of model editing techniques in a more informed manner.

\begin{table*}[ht]
    \centering
    \resizebox{16cm}{!}{
    \begin{tabular}{lll}
    \toprule
        \bf Scheme & \bf Example QA pair & \bf Data source \\
        \midrule
        \multirow{2}{*}{superlative} 
        & Question: What is the largest city in the province of British& \multirow{2}{*}{Existent datasets}\\
        & Columbia, Canada?  \textit{Answer: Vancouver} & \\
        \midrule
        \multirow{2}{*}{sorting} 
        & Question: Sort the following cities based on their city size from small & \multirow{2}{*}{Claude-generation based on superlative}\\
        & to large: Nanaimo, Victoria, Seattle. \textit{Answer: Victoria, Nanaimo, Seattle} & \\
        \midrule
        \multirow{2}{*}{comparative} 
        & Question: What is the name of the distilled spirit that has an alcohol & Existent Dataset,\\
        & content less than or equal to 35.0? \textit{Answer: Mekhong} & Claude-generation based on sorting\\
        \midrule
        \multirow{2}{*}{counting} 
        & Question: How many symphonies were composed by Ludwig van & Manually written,\\
        & Beethoven? \textit{Answer: 9} & Claude-generation\\
        \midrule
        \multirow{2}{*}{aggregation} 
        & Question: How many states/provinces are there in North America,  & \multirow{2}{*}{Claude-generation based on counting}\\
        & i.e. United States, Canada, and Mexico? \textit{Answer: 92} & \\
        \midrule
        \multirow{2}{*}{subtraction} 
        & Question: How many provinces does Mexico have more than Canada? & \multirow{2}{*}{Claude-generation based on counting}\\
        & \textit{Answer: 22} & \\
        \bottomrule
    \end{tabular}
    }
    \caption{Example question-answer pair and data source of each reasoning scheme. Existent datasets include GrailQA \cite{grailQA}, NaturalQuestions \cite{NQ}, ComplexWebQuestions \cite{cwq}, FreebaseQA \cite{freebaseqa}.}
    \label{tab:data_example}
\end{table*}

\section{ReCoE Dataset}\label{dataset}

We employ a hybrid-synthetic approach that combines existing complex QA datasets and LLM-assisted data synthesizing to create the ReCoE dataset.
The dataset is designed to evaluate counterfactual editing across a broad spectrum of reasoning schemes: \textit{superlative}, \textit{comparative}, \textit{sorting}, \textit{counting}, \textit{aggregation}, and \textit{subtraction}. 

A typical datapoint within this dataset encapsulates five key components:
\begin{itemize}
    \setlength\itemsep{0em}
    \item $\mathrm{Q}$: Question that corresponds to each of our defined reasoning schemes
    \item $\mathrm{A}$: Answer with aliases
    \item $\mathbb{F}$: Set of facts that supports the answer ($\mathrm{A}$)
    \item $\mathrm{CA}$: Counterfactual answer with aliases
    \item $\mathbb{CF}$: Set of counterfactuals that supports the counterfactual answer ($\mathrm{CA}$)
\end{itemize}
These components allow us to assess knowledge propagation by editing a language model with the set of counterfactuals $\mathbb{CF}$, and testing if the edited model is able to flip its original answer ($\mathrm{A}$) towards the counterfactual answer ($\mathrm{CA}$) through reasoning.

In this section, we provide a comprehensive overview of the dataset and delve into the nuances of its construction methodology.

\subsection{ReCoE: QA Pairs Construction}
\label{basic_situation}

Table \ref{tab:data_example} presents examples of QA pairs for each reasoning scheme and the corresponding data source. Table \ref{tab:dataset_statistics} presents the dataset statistics including the number of examples and atomic facts of $\mathbb{CF}$ for each reasoning scheme.

The construction of the benchmark starts from basic QA pairs. QA pairs of all superlative data and part of comparative data are from existent QA datasets \cite{grailQA, NQ, cwq, freebaseqa}. QA pairs of counting data are partially hand-written. Based on these datapoints, we create QA pairs for sorting questions, more comparative questions, aggregation questions and subtraction questions synthetically.



\paragraph{Sorting} questions are constructed by prompting Claude based on superlative questions with the following 4 steps. Taking the superlative QA pair in Table \ref{tab:data_example} as an example:\\
\textbf{Step 1:} Prompt Claude to generate the aspect that the sentence is discussing.
\begin{formal}
\small
\textbf{aspect}: city population or city size
\end{formal}
\hspace{-1em}\textbf{Step 2:} Given the QA pair, generate 10 similar or relevant entities to the entity/subject of the question.
\begin{formal}
\small
\textbf{relevant entities}:\\
Large cities in Canada:
Toronto, Montreal, Calgary, Ottawa\\
Cities in British Columbia: 
Victoria, Nanaimo, Nelson\\
Coastal cities in North America: 
Seattle, San Francisco, Los Angeles, Portland
\end{formal}
\hspace{-1em}\textbf{Step 3:} Given the question and generated aspect (Step 1), prompt Claude to generate relevant facts about each generated entity.
\begin{formal}
\small
\textbf{entity facts}:\\
Toronto: a recorded population of 2,794,356 ...\\
Victoria: city size 19.47 km$^2$ (7.52 sq mi) ...\\
Nanaimo: city size 35.25 mi$^2$ ...\\
Seattle: land area of 83.9 square miles (217.3 km$^2$) ...\\
...
\end{formal}
\hspace{-1em}\textbf{Step 4:} Generate 5 sorting questions with the answer using generated entities and their corresponding facts.
\begin{formal}
\small
\textbf{example sorting questions with answers}:\\
\textbf{Question}: Sort the following cities based on their city size from small to large: Nanaimo, Victoria, Seattle.\\
\textbf{Answer}: Victoria, Nanaimo, Seattle.
\end{formal}
Based on the existent superlative sub-dataset, over 10k sorting datapoints can be constructed in the above manner. Since the generated facts and answers based on the facts are not necessarily correct, we leverage \textbf{multi-agent debate} \cite{multiagent} with Claude agents to double-check the generated QA pairs as initial quality control: QA pairs are excluded if the agents, post-debate, converge on different answers to the question. 

\paragraph{Comparative} questions are partially (192) selected from existent datasets. We generate 1,000 more QA pairs with Claude based on generated sorting questions by transforming a sorting question to a comparative question. 

\paragraph{Counting} questions are from 8 different domains: astronomy, book, geography, legal, movie, music, sport, and war. We manually create 5 QA questions for each domain and then prompt Claude to generate more such QA pairs following the examples. Multi-agent debate is again adopted to filter out inaccurate QA pairs. 

\paragraph{Aggregation \& Subtraction} questions are derived from the counting questions and retains the same 8 domains. An aggregation question is formulated by combining two or more counting questions. 
Below is a QA pair example:
\begin{formal}
\small
\textbf{Question}: How many states/provinces are there in North America, i.e. United States, Canada, and Mexico?\\
\textbf{Answer}: 92
\end{formal}
To avoid incongruous or unnatural questions, we employ two strategies:
\begin{enumerate}
    \setlength\itemsep{0em}
    \item Counting questions to be combined are sampled from the same domain.
    \item Filter out unnatural questions with Claude by two criteria: a) whether the question is fluent; b) whether the entities mentioned in the question are compatible in type. For instance, the number of satellites of Earth and that of Mars are compatible for aggregation, while the number of constellations recognized by the International Astronomical Union and the number of Earth's satellites are not. 
\end{enumerate}

\begin{table}[t]
    \centering
    \begin{tabular}{lrr}
    \toprule
    Scheme & Examples & Atomic facts\\
    \midrule
    superlative & 1,172 & 7,248\\
    comparative & 1,171 & 7,218\\
    counting & 643 & 1,309\\
    sorting  & 1,041 & 6,128\\
    aggregation & 509 & 506\\
    subtraction & 501 & 533\\
    \midrule
    Total& 5,037 & 22,942\\
    \bottomrule
    \end{tabular}
    \caption{Statistics of the ReCoE dataset.}
    \label{tab:dataset_statistics}
\end{table}

\subsection{ReCoE: Counterfactual Construction}
After obtaining all the QA pairs for each reasoning scheme, we need to create facts and counterfactual facts to complete each datapoint. The dataset is constructed entirely automatically with the main construction steps illustrated in Figure \ref{fig:construction_pic}.
The construction involves 4 steps:\\
\textbf{Step 1}: After collecting and generating QA pairs, we prompt Claude to create a counterfactual answer. For instance, if the question is about \textit{which team between the Chicago Blackhawks and Pittsburgh Penguins has won more Stanley Cup championships}, and the answer is the \textit{Chicago Blackhawks}, the counterfactual answer would be the \textit{Pittsburgh Penguins}.\\ 
\textbf{Step 2}: Claude is prompted to generate relevant facts about entities mentioned in the answer and counterfactual answer. These facts are then verified for accuracy using a retrieval-augmented method by retrieving relevant paragraphs from Wikipedia using Contriever \cite{izacard2021unsupervised} and corrected if necessary. We filter datapoints to ensure that all QA pairs are valid and (question, counterfactual answer) pairs are invalid. \\
\textbf{Step 3}: To generate counterfactual facts, if a question is superlative, comparative, or sorting (Group 1), we \textbf{swap} the subjects of supporting facts between those related to the actual answer and those pertaining to the counterfactual answer. This process is conducted while eliminating any datapoints that could introduce contradictions in the counterfactual facts generated as a result of this subject swapping; if the question is counting, aggregation, or sorting (Group 2), we \textbf{alter} the facts for the answer to obtain the counterfactual facts for the counterfactual answer while maintaining consistency.\\
\textbf{Step 4}: All sentences in the facts and counterfactual facts are broken down into atomic formats for easier editing. \\

A concrete example in ReCoE is presented in Appendix \ref{sec:example_datpoint}.

\subsection{Fact Representation}
\begin{figure}[htbp]
    \centering
    \includegraphics[width=77mm]{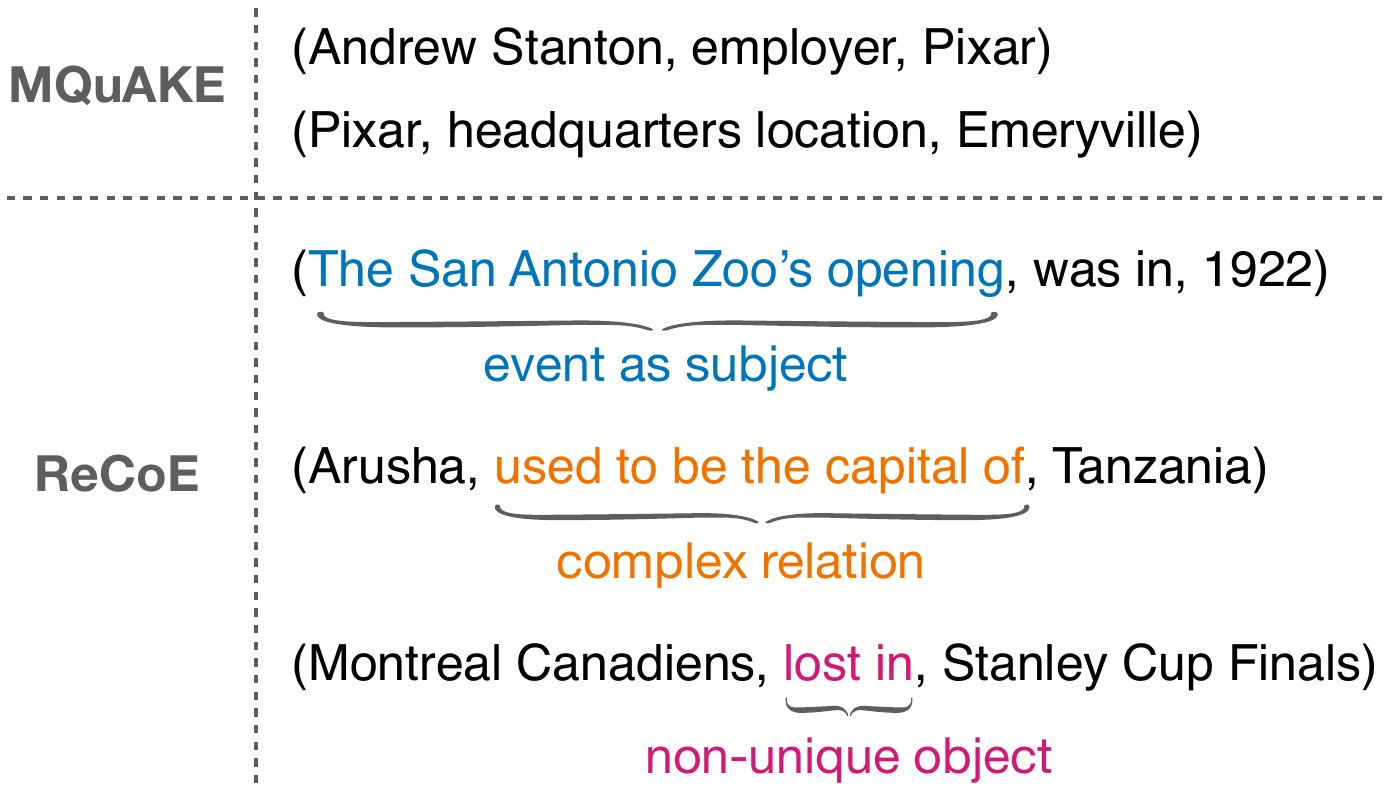}
    \caption{Comparison between fact representations in MQuAKE \cite{MQuAKE} and ReCoE.}
    \label{fig:fact-triples}
\end{figure}
Current benchmarks primarily focus on datasets such as zsRE \cite{entityedit, knowledgeedit, ROME}, COUNTERFACT \cite{ROME}, and MQuAKE \cite{MQuAKE}.
In these datasets, each fact is represented in a clear and unambiguous form as a (subject, relation, object) triplet.
In contrast, our dataset diverges from this norm, featuring facts more commonly encountered in real-world scenarios, typically represented in an OpenIE style.
This style introduces a wider variety of complexities.
As illustrated in Figure \ref{fig:fact-triples}, the atomic facts in ReCoE may involve complex subjects or relations. Moreover, a single relation applied to a subject could correspond to non-unique objects.
These examples highlight the nuanced nature of knowledge updates. We will further discuss the challenges these OpenIE-form fact representations poses to the effectiveness of locate-and-edit methods in a later section (Section \ref{subsec:qlora_vs_memit}).

\begin{table*}[!ht]
    \centering
    \resizebox{16cm}{!}{
    \begin{tabular}{lrrrrrrrr}
        \toprule
        \multirow{2}{*}{\bf Model} & \multirow{2}{*}{\bf Prompt} & \multicolumn{6}{c}{\bf Reasoning Scheme} &\\
        \cmidrule(lr){3-8}
        & & superlative & comparative & counting& sorting& aggregation & subtraction & \it Average\\
        \midrule
        \multirow{2}{*}{7b} & direct & 12.55& 10.93& 31.73& 13.45& 10.22& 10.38 & \it 14.67\\
        & CoT & 28.63& 57.73& 40.90& 10.09& 27.31& 27.94 & \it 32.28\\
        \midrule
        \multirow{2}{*}{13b} & direct & 20.21& 33.13& 33.44& 19.60& 5.30& 11.78 & \it 20.43\\
        & CoT & 30.88& 62.60& 43.23& 16.04& 26.52& 31.54 & \it 35.39\\
        \midrule
        \multirow{2}{*}{33b} & direct & 31.10& 55.42& 46.50& 35.06& 6.48& 11.98 & \it 30.96\\
        & CoT & 37.19& 71.73& 55.37& 36.12& 41.26& 41.52 & \it 46.64\\
        \bottomrule
    \end{tabular}
    }
    \caption{QA accuracy of knowledge probing. Both CoT prompting and model scaling significantly improved the overall performance.}
    \label{tab:model_accuracy}
\end{table*}

\begin{table*}[!ht]
    \centering
    \resizebox{16cm}{!}{
    \begin{tabular}{llrrrrrrrr}
        \toprule
        \multirow{2}{*}{\bf Model}  &\multirow{2}{*}{\bf Editor}& \multirow{2}{*}{\bf Prompt} & \multicolumn{6}{c}{\bf Reasoning Scheme} & \\
        \cmidrule(lr){4-9}
         && & superlative & comparative & counting& sorting& aggregation & subtraction &\it Average\\
        \midrule
        \multirow{6}{*}{7b} &\multirow{2}{*}{InputAug}& direct& 50.30& 41.41& 23.53& 7.86& 7.69& 17.31 & \it 24.68\\
         && CoT & 54.07& 53.99& 34.60& 14.29& 20.86& 10.00 & \it 31.30\\
        \cmidrule(lr){2-10}
  &\multirow{2}{*}{QLoRA}& direct & 0.60& 14.06& 9.80& 11.43& 5.77&7.69 & \it 8.23\\
  && CoT & 3.41& 51.04& 8.37& 7.62& 5.04&8.57 & \it 14.01\\
        \cmidrule(lr){2-10}
  &\multirow{2}{*}{MEMIT}& direct & 0.00& 34.38& 6.37& 3.94& 0.00&7.69 & \it 8.73\\
  && CoT & 0.00& 3.55& 5.32& 3.45& 0.00&4.29 & \it 2.77\\
        \midrule
        \multirow{6}{*}{13b} &\multirow{2}{*}{InputAug}& direct & 59.85& 51.55& 24.19& 15.69& 3.70& 11.86 & \it 27.81\\
         && CoT & 71.78& 66.30&	60.43& 18.56&	25.93& 25.32 & \it 44.72\\
        \cmidrule(lr){2-10}
  &\multirow{2}{*}{QLoRA}& direct & 2.23& 19.07& 5.12& 12.75& 3.70&8.47 & \it 8.56\\
  && CoT & 4.62& 30.83& 8.27& 14.37& 3.70&6.96 & \it 11.46\\
        \cmidrule(lr){2-10}
  &\multirow{2}{*}{MEMIT}& direct & 0.37& 41.49& 11.63& 2.05& 0.00&3.39 & \it 9.82\\
  && CoT & 0.24& 18.69& 7.91& 0.55& 0.74&5.06 & \it 5.53\\
        \midrule
        \multirow{4}{*}{33b} &\multirow{2}{*}{InputAug}& direct & 82.37& 74.88& 23.08& 24.11& 18.18& 11.67 & \it 39.05\\
         && CoT & 73.33& 84.88& 55.90& 32.98& 46.67& 35.10 & \it 54.81\\
        \cmidrule(lr){2-10}
  &\multirow{2}{*}{QLoRA}& direct & 3.14& 12.94& 6.02& 11.78& 3.03&3.33 & \it 6.71\\
  && CoT & 4.24& 24.88& 12.64& 16.76& 5.24&11.06 & \it 12.47\\
        \bottomrule
    \end{tabular}
    }
    \caption{QA accuracy of each reasoning scheme, with different editors, model sizes, and prompting strategies. MEMIT was not implemented on 33b models due to GPU memory constraints. InputAug (upper bound) shows overall reasonable performance, with consistent benefits from CoT prompting and model scaling. Both QLoRA and MEMIT significantly underperform compared to InputAug, with MEMIT showing particularly low performance in certain reasoning schemes like superlative and aggregation. While QLoRA exhibits some improvement from CoT prompting, MEMIT's performance remains consistently poor across all scenarios.}
    \label{tab:main_results}
\end{table*}

\section{Experiment}\label{experiment}
\subsection{Language Models}
We utilize the Tülu series~\cite{tulu} as the base language models to assess knowledge editing approaches. Among various options available, we find Tülu a good candidate for our study for the following reasons: 1) Tülu models are instruction-tuned and provide well-structured responses to user instructions, simplifying our factual evaluation process;\footnote{Their pretrained counterparts, i.e., Llama, however, would require careful output parsing for obtaining answers to ensure reliable evaluation.} 2) The Tülu series includes models of varying sizes, enabling us to explore the impact of model scaling on effective knowledge editing.

\subsection{Knowledge Editing Methods}
We evaluated the following three representative knowledge editing methods\footnote{We do not evaluate meta-learning based methods such as MEND because currently, these methods are not for massive editing and editing massive knowledge leads to low efficacy.}.
\paragraph{Input-augmentation} is an inference-time editing method that appends the counterfactual facts to the question as part of the prompt. Therefore, it does not modify the model weights, but relies on model's capability to perform reasoning from explicit context. It is considered as an upper bound \cite{entityedit,onoe-etal-2022-entity} for model editing.
\paragraph{Finetuning (QLoRA)} performs gradient descent on the new facts to update model parameters. As we are tuning models up to 33 billion parameters, we adopt the parameter-efficient finetuning method QLoRA \cite{qlora} for the sake of computational and time efficiency.
\paragraph{MEMIT} first localizes the factual knowledge in a range of layers in the transformer architecture and then updates the feedforward modules in the layers to insert a massive amount of new facts in the form of triplets. We used the implementation from \citet{zhang2024comprehensive}.

\subsection{Experiment Setting}\label{evaluation}
\paragraph{Factual knowledge probing} We employ both direct prompting and CoT prompting strategies to probe model's proficiency of factual knowledge mastery and reasoning using the ReCoE dataset.
The objective is to ensure that the dataset presents a balanced level of difficulty – neither overly challenging nor too simplistic. This balance is crucial so that the language model under investigation achieves an acceptable level of accuracy for conducting meaningful counterfactual editing experiments, where we observe the model's transition from correct to counterfactual responses. Moreover, the dataset needs to present a degree of challenge to make it a valuable asset for further research on more advanced language models.

\paragraph{Knowledge Editing} We evaluate model's QA performance on the (question, counterfactual answer) pairs of each reasoning scheme post-editing. We use the \textit{correct\_flip} as the primary metric, which measures the percentage of model's predictions that correctly transition from the original answer to the counterfactual answer.\footnote{\textit{correct\_flip} mainly measures the efficacy of \textit{knowledge update}, rather than \textit{knowledge insert}. We focus on the ``update" setting in this work to facilitate our later analysis of the changes in reasoning capabilities post-editing.}

\begin{figure*}[!ht]
    \centering
    \includegraphics[width=\textwidth]{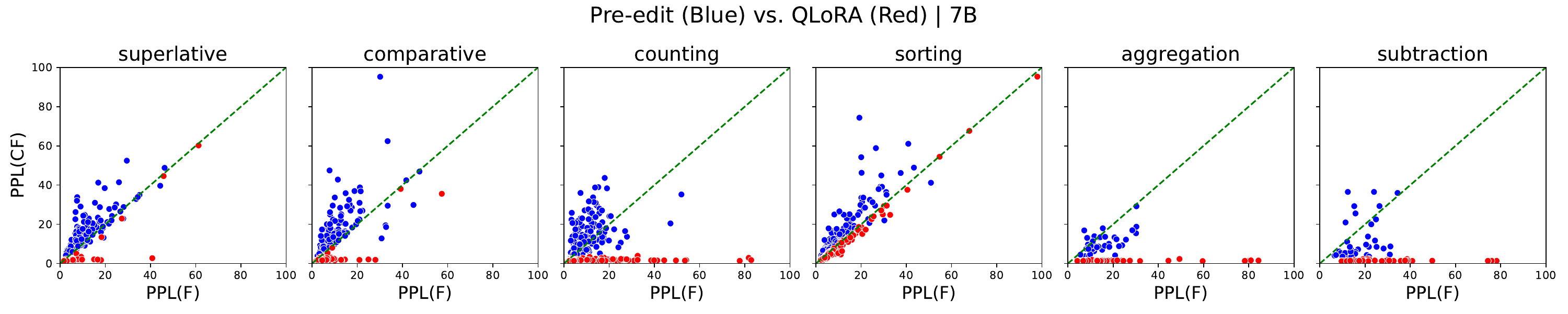}
    \includegraphics[width=\textwidth]{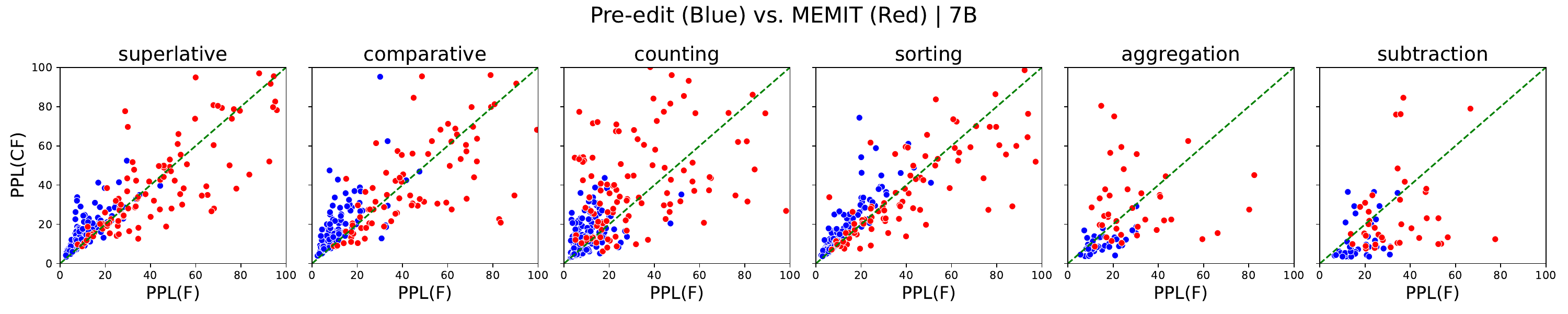}
    \caption{Comparison on fact-wise perplexity over facts ($\mathbb{F}$) and counterfactuals ($\mathbb{CF}$) before and after editing using QLoRA and MEMIT (7B). A successful fact-wise edit is indicated by a transition from the upper-triangle region where $\mathtt{PPL}(\mathbb{CF})>\mathtt{PPL}(\mathbb{F})$ to the lower-triangle region. QLoRA-based finetuning demonstrate notable effectiveness, in contrast to MEMIT. Similar trends are observed with 13b and 33b models, as detailed in Appendix \ref{app:fact-ppl}.}
    \label{fig:factwise_perplexity}
\end{figure*}

\subsection{Experiment Result}
Table \ref{tab:model_accuracy} displays the outcomes of the knowledge probing exercise. The results clearly demonstrate the significant impact of model scaling and the beneficial role of CoT. Table \ref{tab:main_results} summarizes the \textit{correct\_flip} results of InputAug (input-augmentation), QLoRA-based finetuning, and MEMIT.
InputAug involves incorporating counterfactual information into the context, where both model scaling and CoT are shown to be beneficial. Input augmentation is often treated as the upper bound for model editing. But we can see that the performance for aggregation and subtraction is still unsatisfying, below 50\%.
Both QLoRA and MEMIT editing significantly underperform InputAug across all model scales and prompting strategies, indicating failed knowledge propagation of these methods. 
Interestingly, QLoRA-based finetuning, despite its deteriorating performance, can still benefit from CoT prompting and model scaling. In contrast, MEMIT consistently failed, indicating a significant deterioration in the model's reasoning capability.

\section{Analysis}
To comprehensively evaluate how certain knowledge editing methodologies impact model's capability that leads to ineffectiveness in knowledge propagation, our analysis encompasses three key dimensions: \textit{fact-wise editing effectiveness}, \textit{fact recall accuracy}, and \textit{logical coherence} in model's generation.
We assume the reasoning process follows a retrieve-and-generate regime. Formally,
\begin{equation*}
\small
P'(\mathrm{CA}|\mathrm{Q}) = \underbrace{P'(\mathbb{CF}|\mathrm{Q})}_\text{fact recall} \cdot \underbrace{P'(\mathrm{CA}|\mathrm{Q}, \mathbb{CF})}_\text{coherent generation}
\end{equation*}
where $P'$ is the edited LM.
The fact recall component requires 1) each fact within $\mathbb{CF}$ to be effectively edited; and 2) edited model is able to recall these edited facts through generation.
The coherent generation component further requires the generated answer to be logically coherent with the retrieved facts.

\subsection{Fact-wise Editing Effectiveness}
\label{subsec:fact-wise}
This dimension examines the basic efficacy of editing methods. It assesses whether the applied edits achieve their intended modifications successfully, which constitutes the foundational requirement for any knowledge editing approach. Defining $\mathtt{PPL}(\mathbb{F})$ and $\mathtt{PPL}(\mathbb{CF})$ as the averaged perplexity of the facts and the counterfactuals associated to a $(\mathrm{Q}, \mathrm{A}$) pair, and  $\Delta(\mathbb{CF}, \mathbb{F}) = \mathtt{PPL}(\mathbb{CF})$ - $\mathtt{PPL}(\mathbb{F})$), an effective editing over the facts $\mathbb{F}$ to its counterfactual counterpart $\mathbb{CF}$ can be evaluated by the indicator function: $\mathds{1}[\min(\Delta_{pre}(\mathbb{CF}, \mathbb{F}), 0)>\Delta_{post}(\mathbb{CF}, \mathbb{F})]$. This definition of successful editing stipulates that the perplexity of counterfactual sentences must be lower than that of factual sentences. Furthermore, in cases where the perplexity of counterfactual sentences is already lower before editing, it necessitates an even greater disparity between the two.

Fact-wise editing performed by QLoRA demonstrates a high degree of effectiveness, in contrast to MEMIT. In general, $\mathtt{PPL}(\mathbb{CF})$ is lower than $\mathtt{PPL}(\mathbb{F})$ post-editing but MEMIT has the adverse effect of increasing the overall perplexity within the model. Figure \ref{fig:factwise_perplexity} demonstrates the fact-wise editing effectiveness in 7b model using QLoRA and MEMIT. Detailed results are presented in Table \ref{tab:factwise_ppl} in Appendix \ref{app:fact-ppl}.

\subsection{Fact Recall}

\begin{figure*}[t]
\begin{tikzpicture}
    \begin{groupplot}[
        group style={
            group size=2 by 2, 
            vertical sep=1.6cm,
            horizontal sep=1cm
        },
        legend style={at={(1,1.42)},anchor=north,legend columns=-1,font=\small},
        ybar,
        ymin=0,
        ymax=100,
        enlargelimits=0.1,
        symbolic x coords={superlative, comparative, counting, sorting, aggregation, subtraction},
        x tick label style={font=\small, rotate=15, anchor=center, yshift=-8pt},
        xtick=data,
        every node near coord/.append style={font=\tiny},
        nodes near coords align={vertical},
        ylabel near ticks,
        ylabel style={font=\tiny},
        width=8.5cm,
        height=5cm
    ]

    \nextgroupplot[title=Relatedness (7b),bar width=5pt,title style={font=\small}]
    \addplot coordinates {(superlative, 52.79) (comparative, 85.43) (counting, 99.14) (sorting, 78.7) (aggregation, 62.61) (subtraction, 72.29)};
    \addplot coordinates {(superlative, 53.88) (comparative, 79.16) (counting, 97.72) (sorting, 75.51) (aggregation, 86.4) (subtraction, 88.81)};
    \addplot coordinates {(superlative, 39.62) (comparative, 40.08) (counting, 96.96) (sorting, 57.57) (aggregation, 79.13) (subtraction, 87.02)};
    \legend{Pre-Edit,QLoRA,MEMIT}

    \nextgroupplot[title=Consistency (7b),bar width=5pt,title style={font=\small}]
    \addplot coordinates {(superlative, 80.58) (comparative, 65.27) (counting, 43.92) (sorting, 60.5) (aggregation, 68.82) (subtraction, 70.67)};
    \addplot coordinates {(superlative, 28.57) (comparative, 33.39) (counting, 9.89) (sorting, 46.67) (aggregation, 14.63) (subtraction, 15)};
    \addplot coordinates {(superlative, 11.29) (comparative, 8.14) (counting, 4.56) (sorting, 53.04) (aggregation, 8.7) (subtraction, 7.86)};

    \nextgroupplot[title=Relatedness (13b),bar width=5pt,title style={font=\small}]
    \addplot coordinates {(superlative, 53.67) (comparative, 83.54) (counting, 99.64) (sorting, 77.74) (aggregation, 70.71) (subtraction, 77.57)};
    \addplot coordinates {(superlative, 55.12) (comparative, 79.74) (counting, 98.02) (sorting, 78.05) (aggregation, 87.65) (subtraction, 90.93)};
    \addplot coordinates {(superlative, 40.42) (comparative, 45.66) (counting, 97.08) (sorting, 59.7) (aggregation, 80.85) (subtraction, 87.24)};

    \nextgroupplot[title=Consistency (13b),bar width=5pt,title style={font=\small}]
    \addplot coordinates {(superlative, 81.89) (comparative, 63.89) (counting, 48.2) (sorting, 63.73) (aggregation, 76.91) (subtraction, 73.5)};
    \addplot coordinates {(superlative, 28.61) (comparative, 28.49) (counting, 10.43) (sorting, 45.08) (aggregation, 11.85) (subtraction, 24.47)};
    \addplot coordinates {(superlative, 10.75) (comparative, 15.73) (counting, 4.01) (sorting, 46.98) (aggregation, 5.97) (subtraction, 7.59)};

    \end{groupplot}
\end{tikzpicture}
\caption{Fact recall pre- and post-editing: measured by the relatedness and consistency of the decomposed atomic facts in CoT generation against the edited counterfactual facts. While both QLoRA and MEMIT maintains a reasonable degree of relatedness (with QLoRA outperforming MEMIT), there is a significant decline in factual consistency of both methods.}
\label{fig:fact_retrieval}
\end{figure*}
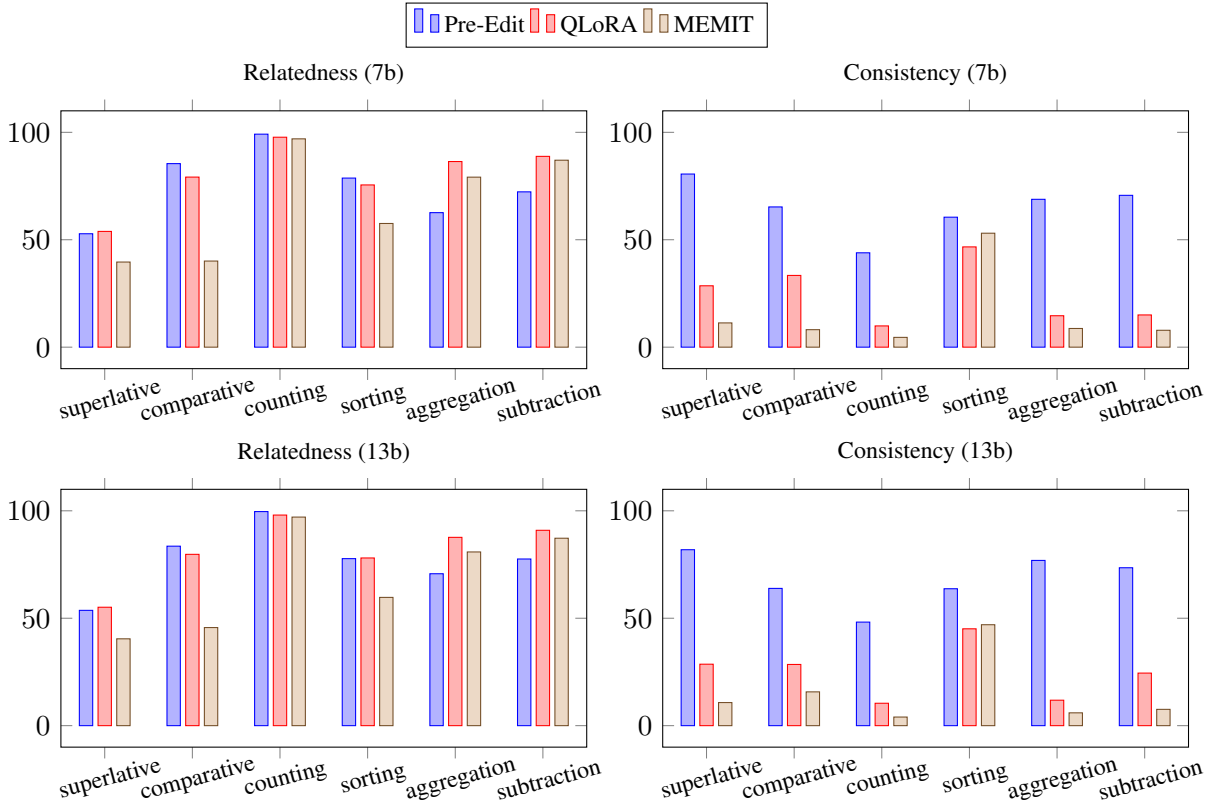

Assuming successful fact-wise editing, we then explore the model's proficiency in recalling and applying these modifications in reasoning tasks. This involves an in-depth analysis of the model's ability to retrieve and utilize relevant information correctly. We focus on evaluating the relatedness and consistency of information within the CoT response against the counterfactual facts. The evaluation metrics are defined as:
\begin{itemize}
    \item \textbf{Relatedness}: this metric assesses how relevant the facts generated by the model are to the designated fact/counterfactual set.
    \item \textbf{Consistency}: this metric calculates the proportion of the model's generated facts that align factually with the fact/counterfactual set.
\end{itemize}

We leverage Claude with dedicated few-shot demonstrations for automatic evaluation. Detailed prompt can be found in Appendix \ref{subsec:fact-recall-eval}.

Results are presented in Figure. \ref{fig:fact_retrieval}. Regarding relatedness, different editing methods show different impacts on the model. In the case of QLoRA, the edited model retains the capability to recall information across different schemes. However, MEMIT shows negative impacts in superlative, comparative, and sorting. In some cases, relatedness post-editing surpasses that of pre-editing. This could be attributed to the introduction of new facts into the model that it previously lacked.

However, both QLoRA-edited and MEMIT-edited models show low consistency results, indicating that they are unable to leverage the edited knowledge in actual use. MEMIT-edited models are worse than QLoRA-edited. It's important to note that this consistency doesn't correlate well with the fact-wise editing effectiveness. This disparity may stem from a lack of generalization during the editing process. Essentially, the model seems to simply memorize the newly edited fact, lacking the ability to extend this understanding to different manifestations of the same concept.

\subsection{Logical Coherence}

\begin{table}[t]
    \centering
    \small{
    \begin{tabular}{lrrrr}
    \toprule
        \bf Scheme & \bf Model & \bf Pre-edit & \bf QLoRA & \bf MEMIT \\
        \midrule
        \multirow{2}{*}{superlative} 
        & 7b & 89.0& 84.2& 8.9\\
        & 13b & 90.1& 85.0& 4.1\\
        \midrule
        \multirow{2}{*}{comparative} 
        & 7b & 73.8& 52.5& 0.7\\
        & 13b & 80.2& 74.1& 4.8\\
        \midrule
        \multirow{2}{*}{counting} 
        & 7b & 91.1& 92.4& 24.0\\
        & 13b & 98.0& 94.1& 32.4\\
        \midrule
        \multirow{2}{*}{sorting} 
        & 7b & 90.5& 85.4& 43.1\\
        & 13b & 90.4& 87.9& 44.3\\
        \midrule
        \multirow{2}{*}{aggregation} 
        & 7b & 92.5& 89.0& 3.6\\
        & 13b & 88.2& 87.2& 5.2\\
        \midrule
        \multirow{2}{*}{subtraction} 
        & 7b & 87.2& 84.8& 10.0\\
        & 13b & 91.4& 92.6& 1.3\\
        \midrule
        \multirow{2}{*}{\it Average} 
        & 7b & 87.4& 81.4& 15.1\\
        & 13b & 89.7& 86.8& 15.4\\
        \bottomrule
    \end{tabular}
    }
    \caption{Coherence of post-editing chain-of-thought generations: percentage of coherent CoT responses among all examples. Coherence is determined by whether the final answer is logically supported by the recalled facts in CoT.}
    \label{tab:cot_coherence}
\end{table}


We investigate whether the logical reasoning capacity of the model is negatively impacted. This is gauged by the coherence of the generated CoT response, specifically evaluating whether the inferred evidence and thought process adequately support the final answer.

Table \ref{tab:cot_coherence} presents results that show a discernible, albeit not substantial, decrease in the QLoRA-edited models. However, there is a surprisingly significant decline in MEMIT-edited models. This indicates a substantial loss of fundamental language modeling abilities, likely due to the catastrophic forgetting associated with MEMIT editing.

\subsection{Discussion}

\subsubsection{QLoRA vs. MEMIT}
\label{subsec:qlora_vs_memit}
For QLoRA, we observe that while it adequately supports fact-wise editing and generally preserves logical coherence, its primary deficiency lies in the retrieval of edited facts. In LLM, the elicitation of knowledge depends heavily on appropriate prompting techniques while our approach involves merely fine-tuning LLMs with atomic fact sentences. Consequently, a potential future direction may involve enriching these atomic facts with more comprehensive contexts prior to their utilization in fine-tuning, as it should enable the model to accurately recall information in response to a diverse set of prompts.

In contrast, the MEMIT model exhibits a decline in all three assessed abilities: fact-wise editing, fact recall, and coherence. Given that our dataset comprises non-synthetic reasoning questions, which often include complex subjects (e.g., an event) and relations, non-unique objects (Figure \ref{fig:fact-triples}), the underperformance of MEMIT suggests its current inadequacy in handling real-world factual knowledge. Notably, MEMIT's most pronounced deficiencies lie in its ability to recall facts and, critically, in maintaining coherence. This observation highlights that the functionalities of edited neurons extend beyond mere fact storage, challenging the assertions made in previous studies \cite{dai2021knowledge, ROME, MEMIT}.

\subsubsection{Effect of Model Scaling}

\begin{figure}[t]
\begin{tikzpicture}
\begin{axis}[
    ybar,
    enlargelimits=0.15,
    legend style={at={(0.95,0.95)},
      anchor=north east,legend columns=1},
    symbolic x coords={KnowledgeProbe,InputAug,QLoRA,MEMIT},
    xtick=data,
    x tick label style={font=\small, rotate=10, anchor=center, yshift=-8pt},
    nodes near coords,
    nodes near coords align={vertical},
    every node near coord/.append style={font=\tiny},
    ylabel near ticks,
    ylabel style={font=\small},
    ]
\addplot coordinates {(KnowledgeProbe,32.28) (InputAug,31.3) (QLoRA,14.01) (MEMIT,2.77)};
\addplot coordinates {(KnowledgeProbe,35.39) (InputAug,44.72) (QLoRA,11.46) (MEMIT,5.53)};
\addplot coordinates {(KnowledgeProbe,46.64) (InputAug,54.81) (QLoRA,12.47)};
\legend{7b,13b,33b}
\end{axis}
\end{tikzpicture}
\caption{Effect of model scaling. The metric (y-axis) refers to averaged \textit{accuracy} over all reasoning schemes for KnowledgeProb, and \textit{correct\_flip} for editing approaches.}
\label{fig:comparison}
\end{figure}
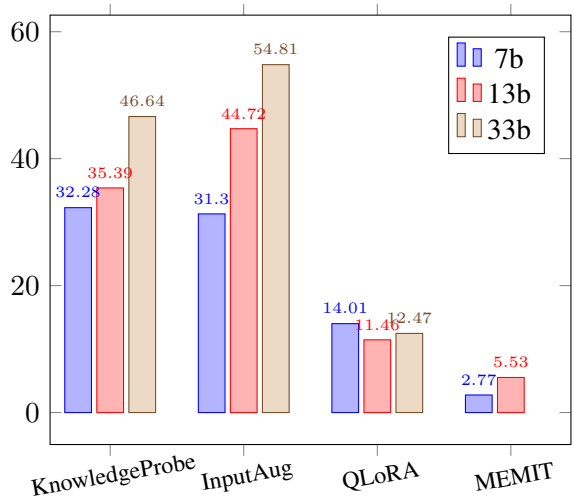

The impact of model scaling is a critical factor in both original knowledge probing and the input-augmentation approach, which are shown in Figure \ref{fig:comparison}, echoes with current studies that larger models inherently possess a more extensive knowledge base and demonstrate superior reasoning capabilities. 

However, experiments in this research reveal that upon editing new knowledge into these models, the size of the model does not correspond to enhanced performance in several dimensions, as shown in Figure \ref{fig:fact_retrieval} and Table \ref{tab:cot_coherence}. Specifically, larger models do not exhibit (1) increased factual effectiveness, (2) improved ability in retrieving facts during chain-of-thought processes in terms of relatedness and consistency, and (3) more coherent chain-of-thought performance. In summary, during the model editing phase, the size of the model does not inherently confer any advantageous properties. Consequently, we have not detected any notable improvements attributable to model scaling, such as facilitation of the editing process or provision of inherent advantages.



\section{Conclusion}\label{conclusion}
In this study, we have developed a novel benchmark, ReCoE, which leverages counterfactual reasoning and is grounded in non-synthetic data for evaluating model editing. Our analysis reveals significant challenges in existing knowledge editing approaches, particularly in their ability to effectively propagate new facts for coherent reasoning. Through this investigation, we have identified key areas where these methods falter. Our work provides a clear direction for future research in this field, aiming to enhance the efficacy and reliability of knowledge editing in computational models.

\bibliography{anthology,custom}


\appendix

\section{Appendix}
\label{sec:appendix}

\subsection{Example Datapoint}
\label{sec:example_datpoint}
%
%
%
%
\begin{figure*}[htbp]
\begin{lstlisting}[upquote=true,morekeywords={question,answer,counterfactual_answer,facts_per_choice,counterfactuals_per_choice,answer_alias,counterfactual_answer_alias,choice_1_facts,choice_2_facts,choice_1_counterfactuals,choice_2_counterfactuals,unchanged_facts,fact,links,atomic_facts,atomic_triples}]
{
  "question": "Who is the last celebrity Brody Jenner had a romantic relationship with?",
  "answer": "Lauren Conrad",
  "counterfactual_answer": "Heidi Montag",
  "facts_per_choice": {
    "choice_1_facts": [
      {
        "fact": "Lauren Conrad and Brody Jenner briefly dated in 2006.",
        "links": [
          "https://en.wikipedia.org/wiki/Lauren_Conrad"
        ],
        "atomic_facts": [
          "Lauren Conrad and Brody Jenner's dating was brief.",
          "Lauren Conrad and Brody Jenner's brief dating was in 2006."
        ],
        "atomic_triples": [
          "(Lauren Conrad and Brody Jenner's dating; was; brief)",
          "(Lauren Conrad and Brody Jenner's brief dating; was in; 2006)"
        ]
      }
    ],
    "choice_2_facts": [
      {
        "fact": "Heidi Montag was never romantically involved with Brody Jenner.",
        "links": [
          "https://en.wikipedia.org/wiki/Heidi_Montag"
        ],
        "atomic_facts": [
          "Heidi Montag was never romantically involved with Brody Jenner."
        ],
        "atomic_triples": [
          "(Heidi Montag; was never romantically involved with; Brody Jenner)"
        ]
      }
    ]
  },
  "counterfactuals_per_choice": {
    "choice_1_counterfactuals": [
      {
        "fact": "Lauren Conrad was never romantically involved with Brody Jenner.",
        "atomic_facts": [
          "Lauren Conrad was never romantically involved with Brody Jenner."
        ],
        "atomic_triples": [
          "(Lauren Conrad; was never romantically involved with; Brody Jenner)"
        ]
      }
    ],
    "choice_2_counterfactuals": [
      {
        "fact": "Heidi Montag and Brody Jenner briefly dated in 2006.",
        "atomic_facts": [
          "Heidi Montag and Brody Jenner's dating was brief.",
          "Heidi Montag and Brody Jenner's brief dating was in 2006."
        ],
        "atomic_triples": [
          "(Heidi Montag and Brody Jenner's dating; was; brief)",
          "(Heidi Montag and Brody Jenner's brief dating; was in; 2006)"
        ]
      }
    ],
    "unchanged_facts": []
  },
  "answer_alias": [
    "Lauren K. Conrad",
    "Lauren Katherine Conrad",
    "L.C."
  ],
  "counterfactual_answer_alias": [
    "Heidi Pratt",
    "Heidi Blair Montag",
    "Heidi B. Montag"
  ]
}
\end{lstlisting}
\caption{An example from the ReCoE dataset (superlative).}
\label{fig:app-recoe-example}
\end{figure*}
The data contract and a specific example from ReCoE is shown in Figure~\ref{fig:app-recoe-example}.

\subsection{Dataset Construction}
\label{construction_process}
Our dataset is constructed almost completely automatically. In this subsection, we discuss in detail how the dataset is constructed step by step. This is the running example that we adopt for illustration:
\begin{formal}
\small
\textbf{Question}: Which team between the Chicago Blackhawks and Pittsburgh Penguins has won more Stanley Cup championships?
\textbf{Answer}: Chicago Blackhawks
\end{formal}

\paragraph{Counterfactual answer generation} Given a question and answer, prompt Claude to generate counterfactual answers. Since the example is a choice question, then the counterfactual answer must be ``Pittsburgh Penguins''. 
\begin{formal}
\small
\textbf{Counterfactual Answer}: Pittsburgh Penguins
\end{formal}
\hspace{-1em}For questions that are not a yes/no question: if the answer is an entity, then the counterfactual answer will be a similar and comparable entity to the answer; if the answer is an ordered sequence of entities or events, the counterfactual answer will be an order with two entities swapped; if the answer is a number, the counterfactual answer will be a close but different number. 

\paragraph{Fact generation} For each triplet of (question, answer, counterfactual answer), prompt Claude to generate relevant facts mentioned entities in the answer and counterfactual answer. Prompt details can be found in Appendix \ref{app:fact_generation}. If the question is yes/no, generate facts on the two entities being compared. In this example, we prompt Claude on the two teams ``Chicago Blackhawks'' and ``Pittsburgh Penguins'' on their number of Stanley Cup winnings. For this example, the mentioned entities are Chicago Blackhawks and Pittsburgh Penguins; their corresponding facts generated are presented below:
\begin{formal}
\small
\textbf{Facts}\\
\textbf{Chicago Blackhawks}: Chicago Blackhawks has won the Stanley Cup Championship six times, in 1930, 1937, 1961, 2010, 2013 and in 2015.\\
\textbf{Pittsburgh Penguins}: The Penguins have won the Stanley Cup five times (1991, 1992, 2009, 2016, and 2017).
\end{formal}

\paragraph{Fact verification}
Hallucination \cite{hallucination} is a severe problem for large language models. Thus, the facts generated from Claude need further verification with truthful and convincing sources. Towards this end, we leverage the retrieval-augmented method to verify each sentence of the generated facts with the following steps: (1) utilize Google Search API to search relevant Wikipedia pages on the question, answer, counterfactual answer, and each sentence of the generated facts (2) chunk content from all the found Wikipedia pages to paragraphs (3) for each sentence, we leverage Contriever \cite{contriever} model implemented by Huggingface\footnote{https://huggingface.co/facebook/contriever} to retrieve the top-5 most relevant paragraphs (4) we prompt Claude using the sentence together with its top-5 most relevant paragraphs to verify whether the sentence is factually correct and if not, modify it based on the retrieved paragraphs. Prompt details can be found in Appendix \ref{app:fact_verification}.

In this example, the generated sentence for Chicago Blackhawks is wrong in the year of 1930, which should be 1933. 
\begin{formal}
\small
\textbf{Facts}\\
\textbf{Chicago Blackhawks}: Chicago Blackhawks has won the Stanley Cup Championship six times, in \textcolor{red}{1930}, 1937, 1961, 2010, 2013 and in 2015. --> Chicago Blackhawks has won the Stanley Cup Championship six times, in \textcolor{red}{1933}, 1937, 1961, 2010, 2013 and in 2015.\\
\textbf{Pittsburgh Penguins}: The Penguins have won the Stanley Cup five times (1991, 1992, 2009, 2016, and 2017): factually correct
\end{formal}

For reference, we also provide Wikipedia links for each sentence in the generated facts.

\paragraph{Datapoint filtering} To guarantee that the provided answer is correct for the question and the counterfactual answer is indeed ``counterfactual'', given the verified facts, we prompt Claude to determine whether the facts support the answer and negate the counterfactual answer and then filter out datapoints with the wrong or outdated answer. This step is necessary as FreebaseQA, GrailQA, ComplexWebQuestions are all based on the Freebase knowledge graph which is outdated. Prompt details can be found in Appendix \ref{app:supporting_check}.

\begin{formal}
\small
\textbf{Counterfactual Facts}\\
\textbf{Pittsburgh Penguins}: Pittsburgh Penguins has won the Stanley Cup Championship six times, in 1933, 1937, 1961, 2010, 2013 and in 2015.\\
\textbf{Chicago Blackhawks}: Chicago Blackhawks have won the Stanley Cup five times (1991, 1992, 2009, 2016, and 2017): factually correct
\end{formal}

\paragraph{Counterfactual Fact Generation}
To create counterfactual facts, we switch the subjects of the facts. Creating counterfactual facts by swapping subjects of supporting facts can guarantee that the generated counterfactual facts support the counterfactual answer.

\paragraph{Datapoints consistency} 
Notice that the generated counterfactual facts could become contradictory to each other if multiple datapoints happen to involve one same fact to edit. For example, one datapoint requires editing the fact about the Amazon River length to be 4,132 miles:
\begin{formal}
\small
\textbf{Question 1}: Which river is longer, the Amazon River or the Nile River?\\
\textbf{Counterfactual Answer}: the Amazon River\\
\textbf{Counterfactual Facts}: \\
\textcolor{red}{The Amazon River has length of 4,132 miles.} \\
The Nile River has length of 3,977 miles.
\end{formal}
Another datapoint requires editing the fact about the Amazon River length to be 3,395 miles:
\begin{formal}
\small
\textbf{Question 2}: Which river is longer, the Amazon River or the Yellow River?\\
\textbf{Counterfactual Answer}: the Yellow River\\
\textbf{Counterfactual Facts}: \\
\textcolor{red}{The Amazon River has length of 3,395 miles.} \\
The Yellow River has length of 3,977 miles.
\end{formal}
This would be problematic for massive editing. Thus to avoid editing facts to be contradictory to each other, for datapoints with contradictory counterfactual editing facts, we randomly retain one of them and remove the others to guarantee consistency among each sub-dataset.

\paragraph{Atomic format of facts}
As multiple editing methods require a (subject, relation, object) format for editing knowledge, in order to facilitate the application of the dataset, we transform each sentence of the facts and counterfactual facts into atomic facts and then atomic triplets. Prompt details can be found in Appendix \ref{app:atomic_breaking_down}.

Let's take the sentence ``Pittsburgh Penguins has won the Stanley Cup Championship six times, in 1933, 1937, 1961, 2010, 2013 and in 2015'' as an example. The atomic facts are sub-sequences of the sentence. Notice that there is no unique way to break a sentence into atomic facts:
\begin{formal}
\small
\textbf{Atomic Facts}\\
\textbf{Atomic fact 1}: Pittsburgh Penguins has won the Stanley Cup Championship six times.\\
\textbf{Atomic fact 2}: Pittsburgh Penguins won the Stanley Cup in 1933, 1937, 1961, 2010, 2013, and 2015.
\end{formal}

\begin{formal}
\small
\textbf{Atomic Triples}\\
\textbf{Atomic triplet 1}: (Pittsburgh Penguins, has won the Stanley Cup Championship, six times)\\
\textbf{Atomic triplet 2}: (Pittsburgh Penguins, won the Stanley Cup in, 1933, 1937, 1961, 2010, 2013, and 2015)
\end{formal}

\paragraph{Alias generation} In evaluation time, we resort to Exact Match on the answer or the counterfactual answer to evaluate models before editing and after editing. Since the model may not generate the exact surface string provided, providing aliases for the answer and the counterfactual answer is necessary to accurately reflect the model capacity as well as the performance of editing methods. Prompt details can be found in Appendix \ref{app:alias_generation}. In this example, the aliases are:
\begin{formal}
\small
\textbf{answer alias}: Blackhawk Division, Hawks\\
\textbf{counterfactual answer alias}: the Pens, Pens
\end{formal}

\paragraph{Summary}
This appendix section presents the basic dataset statistics and the dataset construction process. In each sub-dataset, the datapoint is an n-ary tuple consisting of (question, answer, counterfactual answer, facts, counterfactual facts, answer alias, counterfactual answer alias), where facts and counterfactual facts are lists of dictionaries with 4 keys: sentence, links, atomic$\_$sentences, atomic$\_$triplets. 

\subsection{Fact Generation}
\label{app:fact_generation}
To prompt Claude to generate facts on relevant entities or events, we adopt few shot learning prompting strategy with multiple examples. To save space, we only use 1 example for illustration. See detailed prompt in Figure \ref{fig:fact_generation}.

\begin{figure*}[t]
\begin{tcolorbox}[title=\textbf{Fact Generation Prompt}]
\textbf{Human:}\\

You are given a yes-no question. Your task is to answer the question by explicitly checking the relevant facts of the entities/events being compared in the question.\\

Here is one example:\\

Question: Is Mexico City more populous than Amsterdam?\\

Relevant facts related to <entity1>Mexico City</entity1>:\\
<facts\_for\_entity1>\\
<fact1>Mexico City has 22.2 million</fact1>\\
</facts\_for\_entity1>\\

Relevant facts related to <entity2>Amsterdam</entity2>:\\
<facts\_for\_entity2>\\
<fact1>Amsterdam has 821,752 in population </fact1>\\
</facts\_for\_entity2>\\

Now generate relevant fact for the following question.\\
\{\{question\}\}\\

\textbf{Assistant:}
\end{tcolorbox}
\caption{Fact generation prompt}
\label{fig:fact_generation}
\end{figure*}

\subsection{Fact Verification}
\label{app:fact_verification}
This prompt is used to do factuality verification for generated facts by Claude based on retrieved Wikipedia paragraphs. We adopt few-shot prompting. In the below example, we only present one example in the prompt for demonstration purpose. See detailed prompt in Figure \ref{fig:fact_verification}.
\begin{figure*}[htbp]
\begin{tcolorbox}[title=\textbf{Fact Verification Prompt}]
\textbf{Human:}\\

You are given a sentence and you need to check whether it is consistent with the provided Wikipedia paragraphs.\\

Your task is to judge whether this sentence is factually consistent and potentially rewrite it.\\

If inconsistent: rewrite the sentence by changing it minimally.\\
If consistent: leave it unchanged.\\

Here are four examples:\\

<example>\\
H:\\
Sentence: The Thermosphere, Ionosphere, Mesosphere Energetics and Dynamics (TIMED) satellite is a NASA mission.\\

Wikipedia paragraph: The TIMED (Thermosphere, Ionosphere, Mesosphere, Energetics and Dynamics) mission is dedicated to study the influences energetics and dynamics of the Sun and humans on the least explored and understood region of Earth's atmosphere: the Mesosphere and Lower Thermosphere / Ionosphere (MLTI). The mission was launched from Vandenberg Air Force Base in California on 7 December 2001 aboard a Delta II rocket launch vehicle. The project is sponsored and managed by NASA, while the spacecraft was designed and assembled by the Applied Physics Laboratory at Johns Hopkins University. The mission has been extended several times, and has now collected data over an entire solar cycle, which helps in its goal to differentiate the Sun's effects on the atmosphere from other effects. TIMED Was Launched Alongside Jason-1.\\

A:\\
<response>\\
Based on the paragraph, the sentence is factually consistent.\\
<factuality>Consistent</factuality>\\
Since it is consistent, we do not need to modify it.\\
</response>\\
</example>\\

Now, check the following sentence:\\
sentence: \{\{sentence\}\}\\

Wikipedia paragraph: \{\{retrieved paragraphs\}\}\\

\textbf{Assistant:}
\end{tcolorbox}
\caption{Fact verification prompt. Only 1-shot example is shown for brevity.}
\label{fig:fact_verification}
\end{figure*}

\subsection{Data Filtering with Entailment Verification}
\label{app:supporting_check}
This prompt is used to check whether verified facts support the answer to the question and invalidate the counterfactual answer to the question. We also adopt few-shot prompting and here, we present only one example in the prompt for demonstration purposes. See detailed prompt in Figure \ref{fig:fact_filtering}.

\begin{figure*}[htbp]
\begin{tcolorbox}[title=\textbf{Data Filtering with Entailment Verification Prompt}]
\textbf{Human:}\\

You need to check whether given facts about multiple entities are consistent with the sentence.\\
Here are a few examples:\\

<example>\\
H:\\
<question>Question: Who is the tallest guitarist?</question>\\
<answer>Marc Colombo</answer>\\
<fact>\\
Facts:\\

Marc Colombo: Marc Colombo is an American football player, not a guitarist.\\

Jimmy Page: Jimmy Page is an English musician who gained international fame for his work in the rock band Led Zeppelin. Jimmy Page has a height of 1.82 m or 5 feet 11.5 inches.\\

</fact>\\

A:\\
<response>\\
The given facts state that Marc Colombo is an American football player, thus he cannot be the tallest guitarist. Thus these facts do not support the given answer to the question.\\
<consistency>No</consistency>\\
</response>\\
</example>\\

Here is a new pair of question-answer and facts. Please decide whether the given facts support the provided sentence:\\
<question>Question: \{\{question\}\}</question>\\
<answer>Answer: \{\{answer\}\}</answer>\\
<fact>\\
Facts:\\
\{\{verified facts\}\}\\
</fact>\\

\textbf{Assistant:}
\end{tcolorbox}
\caption{Data filtering with entailment verification prompt}
\label{fig:fact_filtering}
\end{figure*}

\subsection{Atomic Facts Generation}
\label{app:atomic_breaking_down}
This prompt is used to break down sentences in the facts and counterfactual facts into atomic facts and atomic triplets. We also adopt few-shot prompting strategry and only show one example for demonstration purpose. See detailed prompt in Figure.\ref{fig:atomic_prompt}.
\begin{figure*}[htbp]
\begin{tcolorbox}[title=\textbf{Atomic Facts Generation Prompt}]
\textbf{Human:}\\

Given several sentences, break each of them into atomic facts with salient subject, relation, object. After splitting into atomic facts, we also rewrite the atomic fact to a (subject; relation; object) triple.\\
Here are some examples:\\

<example>\\
H:\\
Sentence 1: The Lunar Atmosphere and Dust Environment Explorer (LADEE) was a NASA lunar exploration and technology demonstration mission.\\

A:\\
<response>\\
For the first sentence: \\

This part of sentence "The Lunar Atmosphere and Dust Environment Explorer (LADEE) was a NASA lunar exploration" can form one atomic fact:\\

The subject is "The Lunar Atmosphere and Dust Environment Explorer (LADEE)";\\
The relation is simply "was";\\
The object is "a NASA lunar exploration".\\
<sentence1\_fact1>The Lunar Atmosphere and Dust Environment Explorer (LADEE) was a NASA lunar exploration</sentence1\_fact1>\\
Into triple:\\
<sentence1\_triple1>(The Lunar Atmosphere and Dust Environment Explorer (LADEE), was, a NASA lunar exploration)</sentence1\_triple1>\\

This part of sentence "The Lunar Atmosphere and Dust Environment Explorer (LADEE) was technology demonstration mission" can form another fact:\\
The subject is "The Lunar Atmosphere and Dust Environment Explorer (LADEE)";\\
The relation is simply "was";\\
The object is "technology demonstration mission".\\
<sentence1\_fact2>The Lunar Atmosphere and Dust Environment Explorer (LADEE) was technology demonstration mission</sentence1\_fact2>\\
Into triple:\\
<sentence1\_triple2>(The Lunar Atmosphere and Dust Environment Explorer (LADEE); was; technology demonstration mission)</sentence1\_triple2>\\
</response>\\
</example>\\

Here is a new list of sentences. Please break each of them down into several facts as above.\\
\{\{sentence\}\}\\

\textbf{Assistant:}
\end{tcolorbox}
\caption{Atomic fact breaking-down prompt}
\label{fig:atomic_prompt}
\end{figure*}

\subsection{Alias Generation}
\label{app:alias_generation}
This prompt is used to generate alias for answer and counterfacutal answer. We also adopt few-shot prompting strategry and only show one example for demonstration purpose. See detailed prompt in Figure \ref{fig:alias_generation}.
\begin{figure*}[htbp]
\begin{tcolorbox}[title=\textbf{Alias Generation Prompt}]
\textbf{Human}:\\

Generate aliases for the given entities.\\
Here are some examples.\\

<example>\\

H:\\
entity 1: Luis Fortuno\\
entity 2: Alejandro Garcia Padilla\\

A:\\
<response>\\

For entity 1 Luis Fortuno:\\
<entity1\_alias1>Luis Guillermo Fortuno Burset</entity1\_alias1>\\
<entity1\_alias2>Luis G. Fortuno</entity1\_alias2>\\
<entity1\_alias3>Luis Fortuno</entity1\_alias3>\\
<entity1\_alias4>Luis G. Fortuno</entity1\_alias4>\\
<entity1\_alias5>Luis Guillermo Fortuno Burset</entity1\_alias5>\\

For entity 2 Alejandro Garcia Padilla:\\
<entity2\_alias1>Alejandro Javier Garcia Padilla</entity2\_alias1>\\
<entity2\_alias2>Garcia Padilla</entity2\_alias2>\\
<entity2\_alias3>Garcia-Padilla</entity2\_alias3>\\
<entity2\_alias4>Alejandro J. Garcia-Padilla</entity2\_alias4>\\

</response>\\

</example>\\

Here are two new entities.\\
Please generate their aliases.\\

entity 1: \{\{answer\}\}\\
entity 2: \{\{counterfactual answer\}\}\\

\textbf{Assistant:}
\end{tcolorbox}
\caption{Alias generation prompt}
\label{fig:alias_generation}
\end{figure*}

\subsection{Fact-wise Perplexity}
\label{app:fact-ppl}
Table \ref{tab:factwise_ppl} summarizes the fact-wise editing accuracy of QLoRA and MEMIT, measured using the metric as described in Section \ref{subsec:fact-wise}.

\begin{table*}[ht]
    \centering
    \resizebox{16cm}{!}{
    \begin{tabular}{llrrrrrrarrra}
    \toprule
    \multirow{2}{*}{\bf Scheme}& \multirow{2}{*}{\bf Model}&  \multicolumn{3}{c}{\bf Pre-edit}&  \multicolumn{4}{c}{\bf QLoRA}&  \multicolumn{4}{c}{\bf MEMIT}\\
    \cmidrule(lr){3-5}\cmidrule(lr){6-9}\cmidrule(lr){10-13}
    &  &  PPL(F)&  PPL(CF)&  $\Delta$&  PPL(F)&  PPL(CF)&  $\Delta$&  ACC&  PPL(F)&  PPL(CF)&  $\Delta$& ACC\\
    \midrule
    \multirow{3}{*}{superlative} &  7b&  12.69&  17.52&  -4.84&  5.19&  2.74&  2.45&  97.60&  172.09&  186.84&  -14.75& 55.09\\
     &  13b&  11.55&  14.31&  -2.76&  6.61&  3.84&  2.77&  97.60&  157.98&  177.21&  -19.23& 51.50\\
         &  33b&  10.51&  15.80&  -5.28&  5.42&  2.65&  2.78&  97.01& -- & -- & -- & --\\
    \midrule
     \multirow{3}{*}{comparative}    &  7b&  12.41&  19.15&  -6.74&  5.61&  2.91&  2.70&  97.66&  272.06&  247.97&  24.09& 52.34\\
         &  13b&  9.89&  16.30&  -6.42&  5.66&  3.07&  2.58&  96.88&  356.36&  534.46&  -178.10& 53.91\\
         &  33b&  11.05&  17.42&  -6.37&  6.14&  2.78&  3.35&  99.22& -- & -- & -- & --\\
    \midrule
     \multirow{3}{*}{counting}    &  7b&  10.50&  14.55&  -4.05&  12.82&  1.70&  11.12&  99.02&  182.11&  246.24&  -64.13& 32.35\\
         &  13b&  7.98&  12.73&  -4.75&  13.18&  1.40&  11.78&  98.04&  164.60&  271.20&  -106.60& 29.41\\
         &  33b&  9.52&  13.84&  -4.32&  18.08&  1.51&  16.58&  98.53& -- & -- & -- & --\\
    \midrule
    \multirow{3}{*}{sorting}    &  7b&  16.84&  23.51&  -6.67&  17.53&  15.93&  1.60&  94.49&  186.02&  125.55&  60.47& 64.57\\
         &  13b&  13.40&  19.46&  -6.06&  20.99&  19.24&  1.75&  96.85&  623.16&  393.86&  229.30& 38.58\\
         &  33b&  14.06&  20.44&  -6.38&  16.54&  14.45&  2.09&  96.85& -- & -- & -- & --\\
    \midrule
    \multirow{3}{*}{aggregation}    &  7b&  16.12&  13.96&  2.16&  30.42&  1.47&  28.95&  100.00&  35.19&  37.56&  -2.37& 32.69\\
         &  13b&  12.56&  13.03&  -0.47&  29.11&  1.26&  27.85&  90.38&  21.27&  24.90&  -3.63& 44.23\\
         &  33b&  16.24&  14.58&  1.66&  25.60&  1.43&  24.17&  90.38& -- & -- & -- & --\\
    \midrule
    \multirow{3}{*}{subtraction}    &  7b&  18.74&  12.84&  5.91&  39.62&  1.48&  38.14&  100.00&  39.16&  47.83&  -8.67& 51.92\\
         &  13b&  17.30&  11.99&  5.31&  24.79&  1.29&  23.50&  94.23&  28.89&  31.68&  -2.79& 53.85\\
         &  33b&  18.60&  12.91&  5.69&  33.24&  1.44&  31.80&  98.08& -- & -- & -- & --\\
    \midrule
     &  7b&  14.55&  16.92&  -2.37&  18.53&  4.37&  14.16&  98.13&  147.77&  148.67&  -0.90& 48.16\\
    &  13b&  12.11&  14.64&  -2.52&  16.72&  5.02&  11.71&  95.66&  225.38&  238.89&  -13.51& 45.25\\
    \multirow{-3}{*}{\it Average}    &  33b&  13.33&  15.83&  -2.50&  17.50&  4.04&  13.46&  96.68& -- &  -- & -- & -- \\
    \bottomrule
    \end{tabular}
    }
    \caption{Fact-wise perplexity over facts ($\mathbb{F}$) and counterfactual facts ($\mathbb{CF}$) with pre-edit and post-edit models.}
    \label{tab:factwise_ppl}
\end{table*}


\subsection{Fact Recall Evaluation}
\label{subsec:fact-recall-eval}
Figure \ref{fig:fact_recall_eval} shows the prompt we used for fact recall evaluation of model generation against the set of counterfactual facts.
\begin{figure*}[htbp]
\begin{tcolorbox}[title=\textbf{Fact Recall Evaluation Prompt}]
\textbf{Human:}\\

Given a paragraph and several facts, evaluate for each fact whether the information contained in the paragraph is consistent with it. For each fact, answer <consistent> or <inconsistent>. If the fact is completely unrelated to the paragraph, then ansewr <unrelated>.\\

Below are a few examples:\\

example 1:\\

Paragraph: Snowdon is 1,085 metres (3,560 ft) high. Ben Nevis is 1,345 metres (4,413 ft) high.\\
Fact:\\ 
<1> The height of the summit as 1,085 m (3,560 ft), making Snowdon the highest mountain in Wales. </1>\\
<2> Ben Nevis is 2,000 meters high. </2>\\
<3> Mount Everest at 29,029 ft (8,848 m) is not only the highest peak in the Himalayas, but the highest peak on the entire planet. </3>\\

Evaluation:\\
<1> The fact is talking about the height of the mountain Snowdon, and the paragraph mentions its height as well, thus the fact is related to the Paragraph. With regard to consistency, the paragraph says Snowdon is 1,085 metres (3,560 ft) high and the fact conveys the same thing, they are consistent. <consistent>. </1>\\

<2> The fact is talking about the height of the mountain Ben Nevis, and the paragraph mentions Ben Nevis's height, thus the fact is related. With regard to consistency, the paragraph says Ben Nevis is 1,345 metres (4,413 ft) high but the fact says it to be 2,000 meters high, which is very different, thus inconsistent. <inconsistent> </2>\\

<3> The fact is talking about the height of Mount Everest while the paragraph does not even mention Mount Everest, thus the fact is unrelated. <unrelated> </3>\\

example 2:\\

Paragraph: Houston is located in the state of Texas. Tampa is located in the state of Florida. Florida is located in the southeastern United States. Texas is located in the central United States.\\
Facts:\\
<1> Houston is in Texas. </1>\\
<2> New York City is in the New York state. </2>\\
<3> Texas is in the middle of United States. </3>\\
<4> Florida is in the southeastern United States </4>\\

Evaluation:\\
<1> The paragraph also mentions that Houston is in Texas, which is also indicated in this fact,thus it's both related and consistent. <consistent>. </1>\\

<2> The paragraph does not mention New York City or New York state. Thus it is not related. <unrelated> </2>\\

<3> The paragraph mentions that Texas is located in the central United States, which is indicated in the fact, thus the fact is consistent with the fact. <consistent> </3>\\

<4> The paragraph mentions that Florida is located in the southeastern United States, which is indicated in the fact, thus the fact is consistent with the fact. <consistent> </4>\\

Here is a paragraph and a list of facts:\\
Paragraph: \{\{paragraph\}\}
Facts:\\
\{\{facts\}\}\\

Please judge if the fact and the paragraph is related. If related, indicate whether the fact is consistent with the paragraph using XML tags: <consistent> or <inconsistent>; If not, use the XML tag <unrelated> to indicate.\\

For each fact, Let's think step by step, following the above 2 examples.\\

\textbf{Evaluation:}\\
\end{tcolorbox}
\caption{Claude-based fact recall evaluation}
\label{fig:fact_recall_eval}
\end{figure*}

\subsection{Coherence Evaluation}
\label{subsec:coherence-eval}
Figure \ref{fig:coherence_eval} shows the prompt for coherence evaluation of model's CoT generation, i.e., whether the final answer is supported by the thought process.
\begin{figure*}[htbp]
\begin{tcolorbox}[title=\textbf{CoT Coherence Evaluation Prompt}]
\textbf{Human}:\\

Given a question, evaluate whether the thoughts support the provided answer for the question. Answer <support> or <not-support>.\\

Below are a few examples:\\

example 1:\\

Question: Is Ben Nevis taller than Snowdon?\\
Thoughts: Snowdon is 1,085 metres (3,560 ft) high. Ben Nevis is 1,345 metres (4,413 ft) high.\\
Answer: No\\
Support or not: Since the thoughts say Snowdon is 1,085 metres (3,560 ft) high. Ben Nevis is 1,345 metres (4,413 ft) high, then it means that Ben Nevis is taller than Snowdon. So the provided answer to the question is not supported by the thought. <not-support>\\

example 2:\\

Question: Is Ben Nevis taller than Snowdon?\\
Thoughts: Snowdon is 1,085 metres (3,560 ft) high. Ben Nevis is 1,345 metres (4,413 ft) high.\\
Answer: Yes\\
Support or not: Since the thoughts say Snowdon is 1,085 metres (3,560 ft) high. Ben Nevis is 1,345 metres (4,413 ft) high, then it means that Ben Nevis is taller than Snowdon. So the provided answer to the question is indeed supported by the thought. <support>\\

example 3:\\

Question: Is Houston located more west than Tampa?\\
Thoughts: Houston is located in the state of Texas. Tampa is located in the state of Florida. Florida is located in the southeastern United States. Texas is located in the central United States.\\
Answer: No\\
Support or not: Since the thoughts say Houston is in Texas and Texas in central US, while Tampa is in Florida and Florida is in southeastern US, then Texas is more west to Florida and thus Houston more west than Tampa. The provided answer is thus not supported by the thoughts. <not-support>\\

example 4:\\

Question: Is Houston located more west than Tampa?\\
Thoughts: Houston is located in the state of Texas. Tampa is located in the state of Florida. Florida is located in the southeastern United States. Texas is located in the central United States.\\
Answer: Yes\\
Support or not: Since the thoughts say Houston is in Texas and Texas in central US, while Tampa is in Florida and Florida is in southeastern US, then Texas is more west to Florida and thus Houston more west than Tampa. The provided answer is thus indeed supported by the thoughts. <support>\\

Here is a triple of new question, thought, answer:\\
Question: \{\{question\}\}\\
Thoughts: \{\{thoughts\}\}\\
Answer: \{\{answer\}\}\\

Please judge if the provided answer is supported using <support> or <not-support> to indicate.\\

\textbf{Assistant:}
\end{tcolorbox}
\caption{Claude-based CoT coherence evaluation}
\label{fig:coherence_eval}
\end{figure*}

\end{document}